\documentclass[conference,10pt,a4paper]{IEEEtran}
\IEEEoverridecommandlockouts
\usepackage{cite}
\usepackage{amsmath,amssymb,amsfonts}
\usepackage{algorithmic}
\usepackage{graphicx}
\usepackage{textcomp}
\usepackage{xcolor}
\usepackage{multirow}
\def\BibTeX{{\rm B\kern-.05em{\sc i\kern-.025em b}\kern-.08em
    T\kern-.1667em\lower.7ex\hbox{E}\kern-.125emX}}
\begin{document}

\title{Access Control of Object Detection Models \\ Using Encrypted Feature Maps\\
}

\author{\IEEEauthorblockN{1\textsuperscript{st} Teru Nagamori}
\IEEEauthorblockA{Tokyo Metropolitan University\\
Tokyo,Japan \\
nagamori-teru@ed.tmu.ac.jp}
\and
\IEEEauthorblockN{2\textsuperscript{nd} Hiroki Ito}
\IEEEauthorblockA{Tokyo Metropolitan University\\
Tokyo,Japan  \\
ito-hiroki2@ed.tmu.ac.jp}
\and 
\IEEEauthorblockN{3\textsuperscript{rd} MaungMaung AprilPyone}
\IEEEauthorblockA{Tokyo Metropolitan University\\
Tokyo,Japan \\
april-pyone-maung-maung@ed.tmu.ac.jp}
\and
\IEEEauthorblockN{4\textsuperscript{th} Hitoshi Kiya}
\IEEEauthorblockA{Tokyo Metropolitan University\\
Tokyo,Japan \\
kiya@tmu.ac.jp}
\and

}

\maketitle

\begin{abstract}
In this paper, we propose an access control method for object detection models. The use of encrypted images or encrypted feature maps has been demonstrated to be effective in access control of models from unauthorized access. However, the effectiveness of the approach has been confirmed in only image classification models and semantic segmentation models, but not in object detection models. In this paper, the use of encrypted feature maps is shown to be effective in access control of object detection models for the first time.
\end{abstract}


\section{Introduction}
Deep neural networks (DNNs) and convolutional neural networks (CNNs) have been used widely in various applications such as image classification, semantic segmentation, and object detection\cite{krizhevsky2017imagenet,liu2019recent, galvez2018object}. Training high-performance models is not an easy task, because it requires a large amount of data, powerful computational resources (GPUs), and efficient algorithms. Considering the expertise, cost, and time required for training models, they are considered as a kind of intellectual property that should be protected. \par
There are two approaches to intellectual property protection of models: ownership verification and access control. The difference between these two approaches is that the former aims to identify the ownership of the models, but the latter aims to protect models from unauthorized access\cite{kiya2022overview}.
The ownership verification methods are inspired by watermarking, where a watermark is embedded in the models and the embedded watermark is used to verify the ownership of the models\cite{uchida2017embedding,zhang2018protecting,darvish2019deepsigns,NEURIPS2019_75455e06,xue2021dnn,maung2021piracy}.
However, ownership verification does not have the ability to restrict the execution of the models. Thus, in principle, attackers can freely exploit the models for their own benefit, or use it in adversarial attacks\cite{szegedy2013intriguing}. Therefore, in this paper, we focus on access control, which aims to prevent models from unauthorized access.\par
A number of access control methods have been proposed as a model protection method. 
By encrypting images or feature maps with a secret key, a stolen model cannot be used to its full capacity without a correct secret key \cite{maungmaung_kiya_2021,maung2021protection,ito2021access}.  However, these  methods  have never been applied to object detection models. In this paper, an access control method with encrypted feature maps is applied to object detection models for the first time, and the effectiveness of the proposed method is confirmed in an experiment. 

\begin{figure}[t]
    \centering
    \includegraphics[bb=0 0 750 357,scale=0.3]{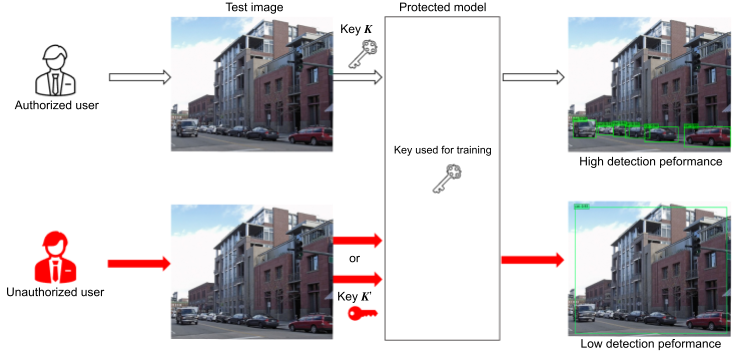}
    \caption{Overview of access control}
    \label{access_control}
\end{figure}

\section{proposed method}
\subsection{Overview}
An overview of the access control to protect the trained models from unauthorized access is shown in Fig. \ref{access_control}. The protected models are trained with the secret key $K$.
When authorized users enter test images and the correct key $K$ into the protected models, the results are equivalent to the models in the unprotected state (the access control is not assumed). In contrast, when unauthorized users without the key $K$ enter only test images or test images and a wrong key $K'$ into the protected models, lower performance results are provided.\par
As access control methods using a secret key, the input image encryption method \cite{maungmaung_kiya_2021} and the feature map encryption method \cite{ito2021access} have been proposed.
Maung's method\cite{maungmaung_kiya_2021} focuses on access control of image classification models, where input images are divided into blocks and encrypted with a secret key using methods such as pixel shuffling, bit flipping, and format-preserving Feistel-based encryption (FFX)\cite{bellare2010addendum}. These encrypted images are used as training and test images. Since this method encrypts the images block by block, it changes the spatial information and cannot be used to protect the object detection models described below. \par
Ito's method \cite{ito2021access} focuses on access control of semantic segmentation models, where models are trained and tested by randomly permuting the channels of feature maps selected by a secret key. This encryption method is spatially invariant. This property was confirmed to be very important for some applications such as semantic segmentation \cite{ito2021access}. Although this method has been validated for semantic segmentation, it has not been validated for object detection models. 
Therefore, in this paper, we propose an access control method for object detection models based on this method.

\begin{figure*}[htb]
    \centering
     \includegraphics[bb=0 0 1327 305,scale=0.35]{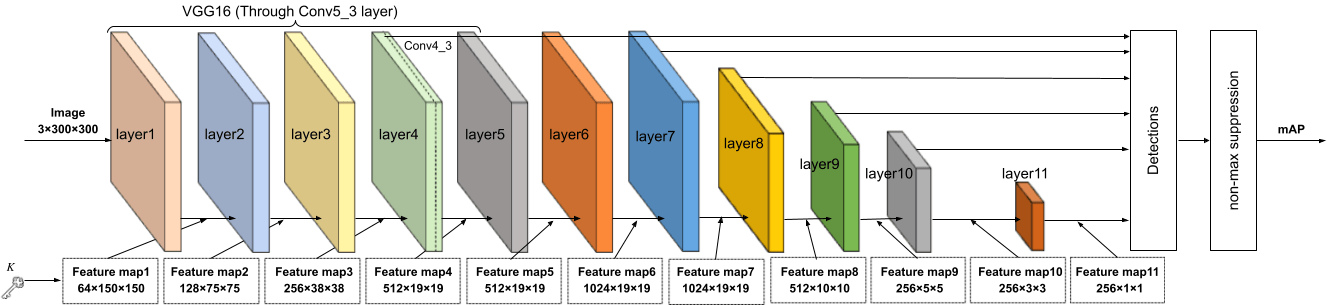}
    \caption{Architecture of object detection model (SSD300)}
    \label{proposed_model}
\end{figure*}

\begin{figure}[tb]
    \centering
    \includegraphics[bb=0 0 699 231,scale=0.35]{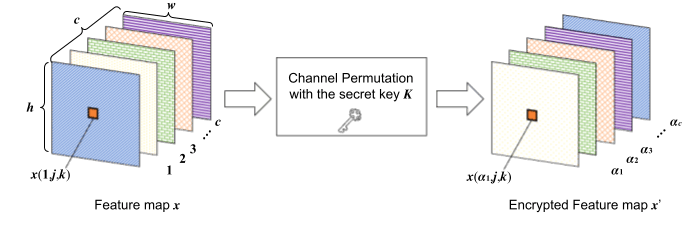}
    \caption{Feature map encryption\cite{ito2021access}}
    \label{transform}
\end{figure}

\subsection{Encryption Method}
There are multiple feature maps in CNNs as shown in Fig. \ref{proposed_model}. In the proposed method, selected feature maps are transformed by using a secret key in accordance with the procedure of learnable image encryption\cite{maungmaung_kiya_2021,adv-def}. Below is the procedure of the encryption, where  $x$ is a selected feature map with a dimension of ($c \times h \times w$), $c$ is the number of channels, $h$ is the height, and $w$ is the width of the feature map.
\begin{itemize}
    \item[1)]Generate a random vector with a size of $c$ using a secret key as in (1).
        \begin{equation}
            [\alpha_1,.,\alpha_i,\alpha_{i'},...,\alpha_c],\alpha_i \in \left\{1,...,c\right\}
        \end{equation}
         where{ $\alpha_i \ne \alpha_{i'}$ }, if $i\ne i'$.\par
    \item[2)]Replace each element $x(i,j,k)$ of $x$, $i\in \left\{1,...,c\right\}, j \in \left\{1,...,h\right\}, k \in \left\{1,...,w\right\}$ with $x(\alpha_i,j,k)$ so that $x$ is transformed into a feature map $x'$. Note that elements of $x'$, $x'(i,j,k)$ is equal to $x(\alpha_i,j,k)$.
\end{itemize}
This encryption is a spatial-invariant operation, so the spatial information of feature maps can be maintained (see Fig. \ref{transform}). This property is very important in object detection tasks, which predict position and classes of objects. 

\subsection{Model Training and Testing}
In the proposed method, the previously mentioned transformation method is applied to selected feature maps in an object detection model at each iteration for a training model.
 SSD300\cite{liu2016ssd} based on VGG16\cite{simonyan2014very}, which was pretrained on the ILSVRC CLS-LOC dataset\cite{russakovsky2015imagenet}  is used as an object detection model in this
paper, where SSD300 has 11 feature maps as illustrated in Fig. \ref{proposed_model}. \par
In testing the trained model, authorized users have the same key that is used for the training. When Authorized users apply query images to the model, they transform the same feature maps that are selected for the training with the key. If unauthorized users without the correct key steal the protected model, we assume that they transform the feature maps with an incorrect key or use the model without the transform.

\subsection{Requirements of Protected Models}
Protected models should meet the following requirements.
\begin{itemize}
    \item It provides almost the same performance as that of models trained with plain images to authorized users with the secret key.
    \item It provides a degraded performance to unauthorized users without the correct
key.
\end{itemize}
\section{experiments and results}
\subsection{Setup}
We used the PASCAL visual object classes (VOC) challenge 2007 \cite{everingham2010pascal}, and 2012 \cite{everingham2015pascal} trainval datasets for training, and the PASCAL VOC 2007 test dataset for testing. 
For data augmentation, the random sample crop, horizontal flip, and some photometric distortions described in \cite{liu2016ssd} were used for training models. In addition, due to the restrictions of SSD300 shown in Fig. \ref{proposed_model}, input images were resized to $300\times300$ pixels.\par
Models were trained by using a stochastic gradient descent (SGD) optimizer with an initial learning rate of $10^{-\:3}$, a momentum value of 0.9, a weight decay value of 0.0005, and a batch size of 32.
Models were also trained with a learning rate of $10^{-\:3}$ for 60k iterations, then continue training for 20k iterations with $10^{-\:4}$ and 40k iterations with $10^{-\:5}$. The overall objective loss function is a weighted sum of the localization loss and the confidence loss. 
In this paper, the confidence loss was the cross-entropy loss over multiple classes confidences, and the localization loss was the Smooth L1 loss between the predicted position and the ground truth position.

\begin{figure*}[t]

\scalebox{0.8}[0.8]{
    
    \begin{tabular*}{50mm}{@{\extracolsep{\fill}}c|c|ccc}
        Ground Truth&Baseline&Correct ($K$)&Plain&Incorrect ($K'$) \\
        \begin{minipage}{4truecm}
             \centering
              \includegraphics[bb=0 0 559 453,scale=0.18]{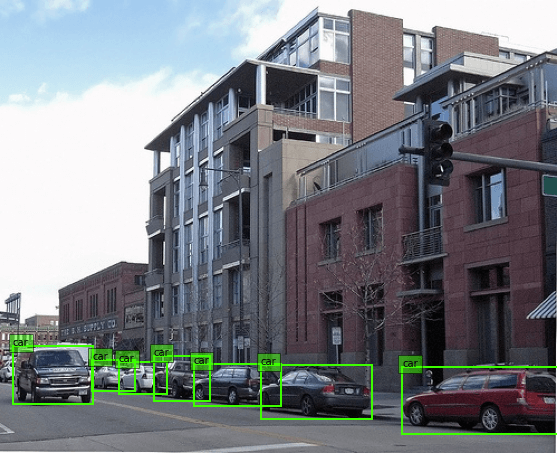}
            \end{minipage}
        &
        \begin{minipage}{4truecm}
             \centering
              \includegraphics[bb=0 0 556 453,scale=0.18]{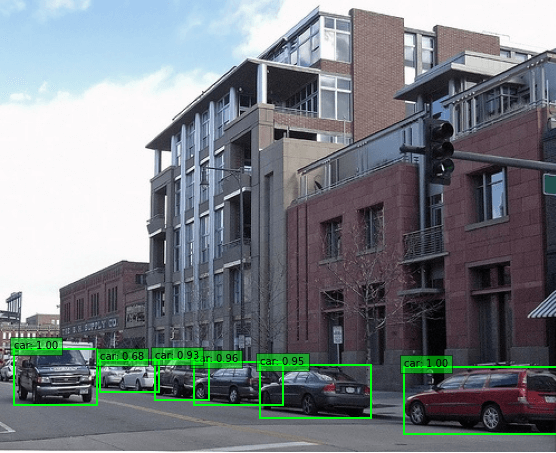}
            \end{minipage}
        &
        \begin{minipage}{4truecm}
             \centering
              \includegraphics[bb=0 0 558 452,scale=0.18]{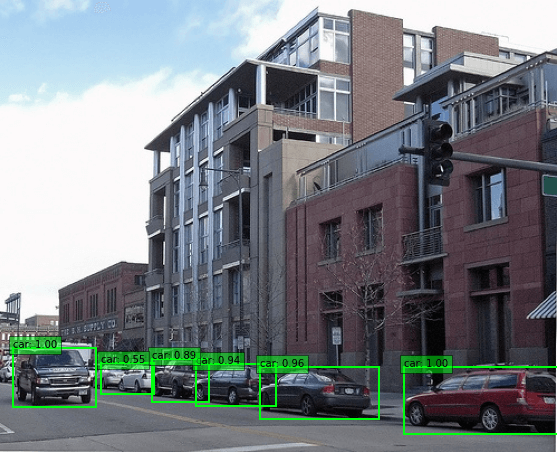}
            \end{minipage}
        &
        \begin{minipage}{4truecm}
             \centering
              \includegraphics[bb=0 0 559 453,scale=0.18]{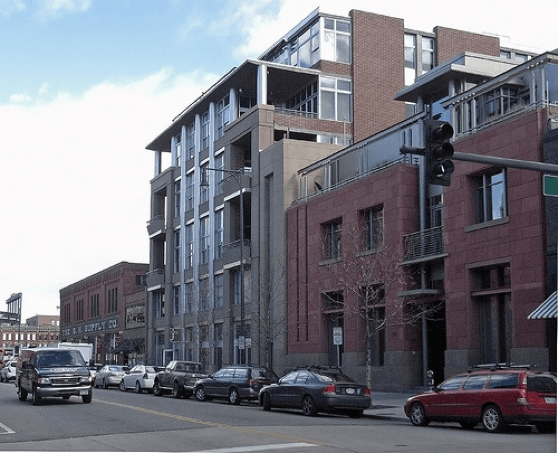}
            \end{minipage}
        &
        \begin{minipage}{4truecm}
            \centering
              \includegraphics[bb=0 0 558 452,scale=0.18]{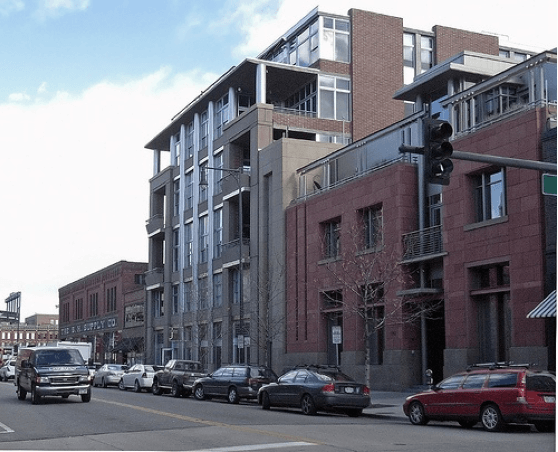}
            \end{minipage}\\
        \begin{minipage}{4truecm}
             \centering
              \includegraphics[bb=0 0 558 369,scale=0.18]{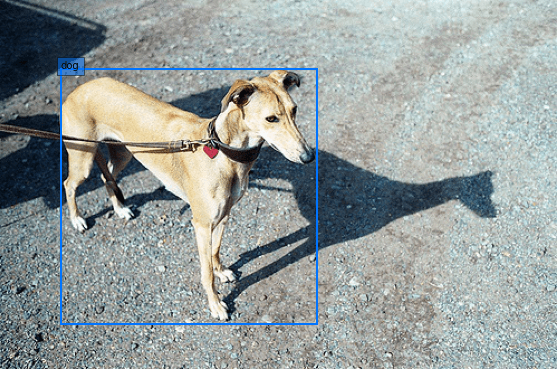}
            \end{minipage}
        &
        \begin{minipage}{4truecm}
             \centering
              \includegraphics[bb=0 0 558 368,scale=0.18]{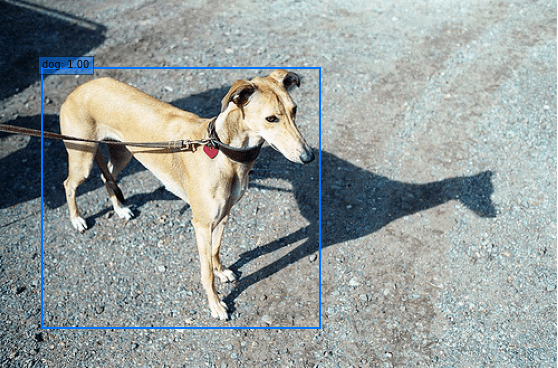}
            \end{minipage}
        &
        \begin{minipage}{4truecm}
             \centering
              \includegraphics[bb=0 0 559 369,scale=0.18]{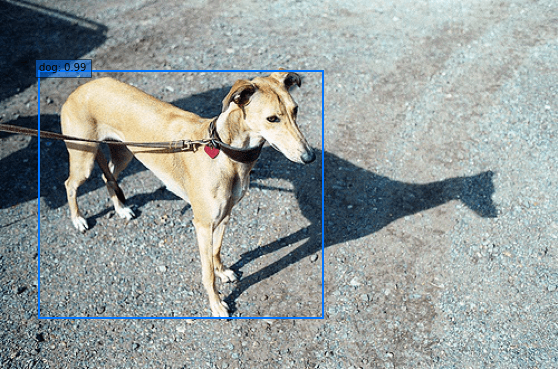}
            \end{minipage}
        &
        \begin{minipage}{4truecm}
             \centering
              \includegraphics[bb=0 0 561 383,scale=0.18]{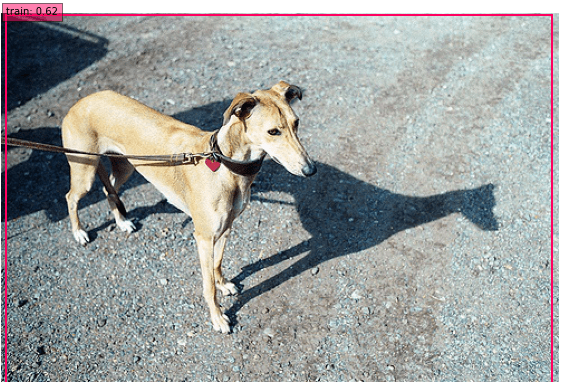}
            \end{minipage}
        &
        \begin{minipage}{4truecm}
             \centering
              \includegraphics[bb=0 0 561 370,scale=0.18]{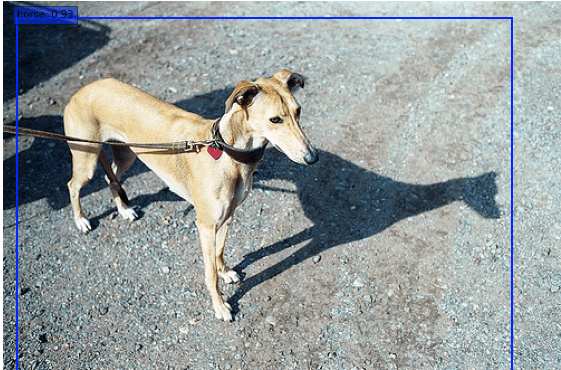}
            \end{minipage}\\
        \begin{minipage}{4truecm}
             \centering
              \includegraphics[bb=0 0 414 546,scale=0.18]{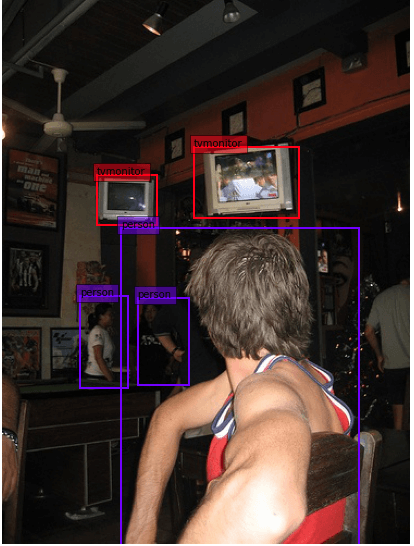}
            \end{minipage}
        &
        \begin{minipage}{4truecm}
             \centering
              \includegraphics[bb=0 0 427 546,scale=0.18]{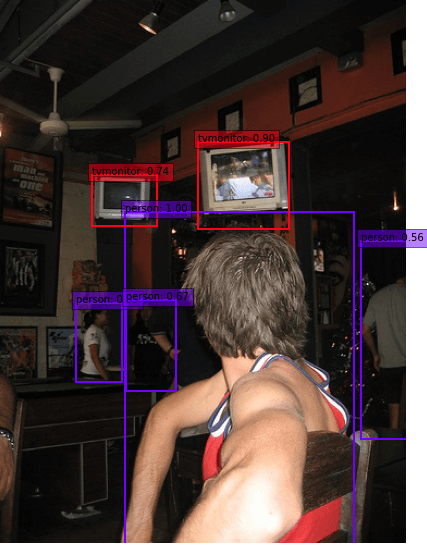}
            \end{minipage}
        &
        \begin{minipage}{4truecm}
              \centering  
              \includegraphics[bb=0 0 405 548,scale=0.18]{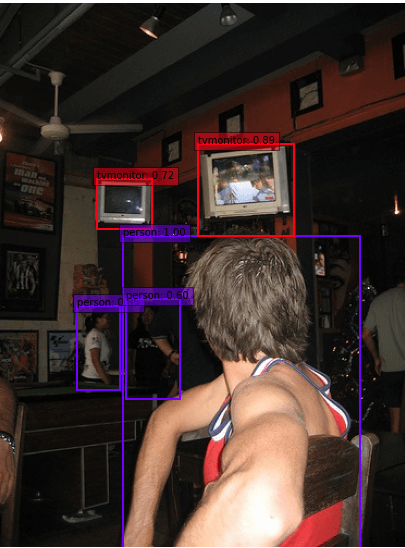}
            \end{minipage}
        &
        \begin{minipage}{4truecm}
             \centering
              \includegraphics[bb=0 0 409 550,scale=0.18]{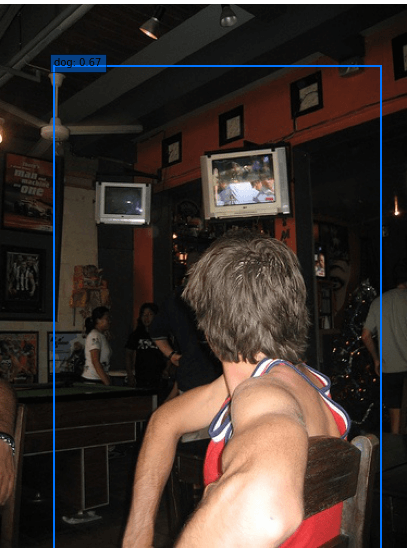}
            \end{minipage}
        &
        \begin{minipage}{4truecm}
            \centering
              \includegraphics[bb=0 0 410 564,scale=0.18]{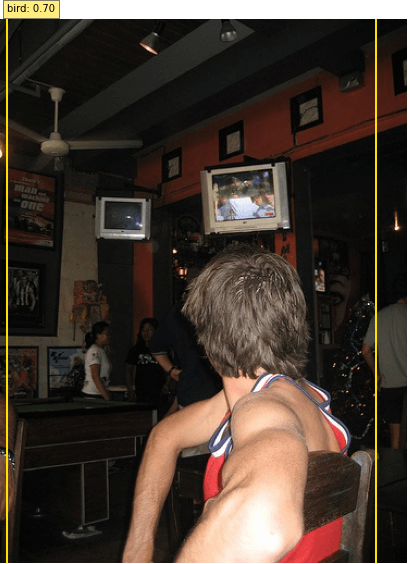}
            \end{minipage}\\
        
    \end{tabular*}
    }
    \caption{Examples of experimental result (Model-4)}
    \label{detection_result}
\end{figure*}

\subsection{Detection Performance}
Mean average precision (mAP) \cite{liu2016ssd} with a range [0,1] was used as a metric for evaluating detection performance, where when a mAP value is closer to 1, it indicates a higher accuracy.
In the experiment, a selected feature map was transformed with a key $K$ in accordance with the procedure in sec. II. In Table \ref{table:result}, Correct ($K$) indicates that the selected feature map was transformed with the same key $K$ as the training. Model-1 means that feature map 1 was selected for encryption, and Baseline indicates that training and testing were performed without any encryption. Fig. \ref{detection_result} shows examples of experimental result where Model-4 was used. \par
From Table \ref{table:result} and Fig. \ref{detection_result}, the proposed method provides the same prediction results as the Baseline when the feature map is transformed using the correct key for the test.

\begin{table}[bt]
 \caption{Detection accuracy (mAP) of proposed models}
 \label{table:result}
 \centering
  \begin{tabular}{|c|ccc|}
   \hline
   Selected feature map  & Correct ($K$) & Plain & Incorrect ($K'$) \\
   \hline
   Model-1& 0.7244 & 0.1363 & 0.0421 \\
   Model-2& \textbf{0.7611} & \textbf{0.0091} & \textbf{0.0180} \\
   Model-3& 0.7475 & 0.0091 & 0.0078 \\
   Model-4& \textbf{0.7611} & \textbf{0.0023} & \textbf{0.0043} \\
   Model-5& \textbf{0.7587} & \textbf{0.1672} & \textbf{0.1624} \\
   Model-6& \textbf{0.7617} & \textbf{0.1732} & \textbf{0.1672} \\
   Model-7& \textbf{0.7695} & \textbf{0.1768} & \textbf{0.1750} \\
   Model-8& 0.7677 & 0.3529 & 0.3415\\
   Model-9& 0.7705 & 0.5767 & 0.5678 \\
   Model-10& 0.7705 & 0.7177 & 0.7027 \\
   Model-11& 0.7512 & 0.7314 & 0.7252 \\
   \hline
   Baseline& \multicolumn{3}{c|}{0.7690}\\
   \hline
  \end{tabular}
\end{table}

\subsection{Robustness against Unauthorized Access}
Two types of unauthorized access were considered in the experiment.
Plain in Table \ref{table:result} represents that an unauthorized user without the key applied query images to protected models, without transforming the selected feature map. Incorrect ($K'$) in Table \ref{table:result} is that an unauthorized user without the key applied query images to protected models, after transforming the selected feature map with a randomly generated key $K'$. 
The result of Incorrect ($K'$) are the average value of 100 tests with random keys. \par 
From the table, Model-1$-$7 provided a low detection accuracy for both Plain and Incorrect ($K'$). On the other hand, when transforming the feature map of a deep layer, the resistance to unauthorized access is lost.
We consider that the reason for this lies in the structure of SSD300.
In order to detect objects of various scales in SSD, detection is performed using features from multiple layers (see Fig \ref{proposed_model}). Therefore, for example, in Model-9, layers 4, 7, and 8 can use the same features as Baseline. In other words, we consider that this is because the number of the same features as Baseline increases in the deeper layers.\par
From Fig. \ref{detection_result}, the detection performance degraded significantly when the model was used illegally. Accordingly, the proposed models were robust enough against the unauthorized access.
\begin{table}[tb]
 \caption{Detection accuracy (mAP) of models \\ with encrypted input images}
 \label{table:SHF}
 \centering
  \scalebox{0.9}{
    \begin{tabular}{|c|c|ccc|}
       \hline
       method & block size  & Correct ($K$) & Plain & Incorrect ($K'$) \\
       \hline
       \multirow{5}{*}{pixel shuffling (SHF)} & 1 & 0.7710 & 0.7598 & 0.7603 \\
       &4& 0.7154 & 0.5745 & 0.3883 \\
       &12&  0.4891 & 0.1976 & 0.0910 \\
       &20& 0.0083 & 0.0086 & 0.0065 \\
       &60& 0.1284 & 0.0480 & 0.0416 \\
       \hline
       \multicolumn{2}{|c|}{Proposed (Model-4)}& \textbf{0.7611} & \textbf{0.0023} & \textbf{0.0043} \\
       \hline
       \multicolumn{2}{|c|}{Baseline} & \multicolumn{3}{c|}{0.7690}\\
       \hline
    \end{tabular}
  }
\end{table}
\subsection{Comparison with encryption of input images}
 The proposed method was compared with a method to protect models with encrypted input images, which was proposed for image classification models\cite{maungmaung_kiya_2021}. In the method, there are three block-wise methods: pixel shuffling, bit flipping, and Format-preserving Feistel-based encryption (FFX)\cite{bellare2010addendum}, for encrypting
input image. \par
In this paper, pixel shuffling (SHF) with a block size of 1, 4, 12, 20, or 60 were
applied to input images, and the encrypted images were used for training and
testing. \par
 The experimental conditions are the same as in  \textit{A} of sec. III. From Table \ref{table:SHF}, the detection accuracy was significantly lower than the proposed method under
almost all block sizes. When the block size was small, the detection accuracy
was high, but the resistance to unauthorized access was also degraded, so the
models were not protected \cite{maungmaung_kiya_2021}. In contrast, when the block size was large, the
resistance to unauthorized access was stronger, but the detection accuracy was greatly degraded. Therefore, the conventional method with encrypted input images is not effective in object detection models. 
 
\section{Conclusion}
We proposed an access control method that uses encrypted feature maps transformation for object detection models for the first time. In the experiment, the proposed access control method was demonstrated not only to provide a high detection accuracy but also to robust enough against two types of unauthorized access.

\section*{Acknowledgement}
This study was partially supported by JSPS KAKENHI (Grant Number JP21H01327).

\bibliographystyle{IEEEtran} 
\bibliography{ref} 
\end{document}